\documentclass[10pt,twocolumn,letterpaper]{article}

\usepackage{template}
\usepackage{times}
\usepackage{epsfig}
\usepackage{graphicx}
\usepackage{amsmath}
\usepackage{amssymb}
\usepackage[font=small]{caption}
\usepackage[labelfont=small]{subcaption}
\usepackage{booktabs, multirow}
\usepackage{enumitem}

\usepackage{glossaries}

\makeatletter
\renewcommand{\subsubsection}[1]{%
\vspace{.5em}\noindent\textbf{#1. }%
\@ifnextchar\par{\@gobble}{}}
\makeatother

\usepackage[pagebackref=true,breaklinks=true,colorlinks,bookmarks=false]{hyperref}
\usepackage{cleveref}
\usepackage{flushend}

\newacronym{cnn}{CNN}{Convolutional Neural Network}
\newacronym{rnn}{RNN}{Recurrent Neural Network}
\newacronym{dfdc}{DFDC}{Deepfake Detection Challenge}
\newacronym{ltpa}{LTPA}{Learn to Pay Attention}
\newacronym{shap}{SHAP}{Shapley Additive Explanation}
\newacronym{gradcam}{GradCAM}{Gradient-weighted Class Activation Mapping}
\newacronym{lime}{LIME}{Local Interpretable Model-agnostic Explanations}

\templatefinalcopy %

\iftemplatefinal\fi

\begin{document}

\title{What's wrong with this video? Comparing Explainers for Deepfake Detection}

\author{Samuele Pino\\
Politecnico di Milano\\
{\tt\small samuele.pino@mail.polimi.it}
\and
Mark James Carman\\
Politecnico di Milano\\
{\tt\small mark.carman@polimi.it}
\and
Paolo Bestagini\\
Politecnico di Milano\\
{\tt\small paolo.bestagini@polimi.it}
}

\maketitle

\begin{abstract}
Deepfakes are computer manipulated videos where the face of an individual has been replaced with that of another. Software for creating such forgeries is easy to use and ever more popular, causing serious threats to personal reputation and public security. The quality of classifiers for detecting deepfakes has improved with the releasing of ever larger datasets, but the understanding of why a particular video has been labelled as fake has not kept pace.

In this work we develop, extend and compare white-box, black-box and model-specific techniques for explaining the labelling of real and fake videos. In particular, we adapt SHAP, GradCAM and self-attention models to the task of explaining the predictions of state-of-the-art detectors based on EfficientNet, trained on the Deepfake Detection Challenge (DFDC) dataset. We compare the obtained explanations, proposing metrics to quantify their visual features and desirable characteristics, and also perform a user survey collecting users' opinions regarding the usefulness of the explainers.

\end{abstract}

\section{Introduction}
\label{sec:introduction}
Deepfakes are a type of video forgery consisting in replacing the face of an original subject in the video with the face of someone else. Deepfakes are generated through a deep learning technique that ``learns'' the look of a subject and then it is able to apply that look to different subjects with low effort \cite{afchar2018mesonet}. Since when deepfake technology emerged on the Internet in the shape of adult content, it has got progressively better, up to the point where today human perception is often fooled into believing the new forged videos to be true.

While there exist positive use cases of this technology in the multimedia industry \cite{Chawla2019DeepfakesH, Gardiner2019FacialRS}, the ability of swapping someone's face poses serious problems. Threats range from easy and effective disinformation spreading by faking public authorities, to online abuse and financial fraud. A major issue is the weakening of videos as evidences in law enforcement, as they cannot be considered reliable anymore.

A recent competition called \gls{dfdc} has been started by Facebook in collaboration with other companies to tackle the deepfake problem. The competition contributed to the research with a new video dataset \cite{dolhansky2020deepfake} and by making public several new deepfake detection systems \cite{bonettini2020video}.

Most of the deepfake detectors are limited to output a prediction on whether the video under analysis is fake or not. Independently of how accurate may this prediction be, it alone does not carry any additional explanations about the reasons for its output. The lack of explainability and prediction insights, typical of deep learning tools like most of deepfake detectors, will eventually hamper their use in practice by making it harder for users to trust them.

In this work we investigate the explainability problem specifically applied to the deepfake detection field. We extend and compare different explanation techniques, proposing metrics to quantify their visual features and desirable characteristics and we show the results of a user study.

The main contributions of the work are the following:
\begin{itemize}[noitemsep]
\item We adapt and extend four different explanation techniques based on black-box (i.e., \acrshort{shap} \cite{NIPS2017_7062}), white-box (i.e., \acrshort{gradcam} \cite{8237336}) and model-based approaches (i.e., \acrshort{ltpa} \cite{jetley2018learn} and Bonettini \cite{bonettini2020video}). Our extensions include: (i) a novel method for segmenting videos for use with black-box techniques, (ii) the first use of the \gls{ltpa} technique with the EfficientNet architecture and (iii) an adaptation of \gls{gradcam} pipeline to generate complete explanations in the binary classification setting.
\item We define new task-specific methods for evaluating deepfake explanation techniques. In particular, we define four different intrinsic metrics (variance, intra-frame consistency, inter-frame consistency and centredness) to compare video explanations without access to ground truth explanations. In addition, we develop an extrinsic evaluation methodology to evaluate the explanations subjectively.
\item We empirically compare the explanation approaches based on the defined intrinsic measures, demonstrating the inherent effectiveness of certain techniques to provide stable and coherent explanations.
\item We perform a user survey to provide a subjective assessment of the perceived quality and usefulness of different visual explanations obtained by the implemented techniques.
\end{itemize}

The rest of the paper is structured as follows.
Section~\ref{sec:related_work} introduces the reader to deepfake detection methods and machine learning explanation techniques.
Section~\ref{sec:approach} provides the details of the investigated explanation methodologies.
Section~\ref{sec:experiments} illustrated our experimental campaign.
Section~\ref{sec:conclusions} concludes the paper.

\section{Related Work} \label{sec:related_work}

The deepfake detection field has seen a high number of contributions in the past two years. While some approaches tend to rely on a specific set of hand-crafted features that can help to discriminate forged videos from real ones \cite{korshunov2018deepfakes, li2018ictu, li2018exposing, yang2018exposing, Matern2019exploiting, Agarwal2019protecting, li2019face, Guarnera_2020_CVPR_Workshops}, others prefer to rely on purely data-driven approaches in a supervised environment \cite{afchar2018mesonet, guera2018deepfake, roessler2019ff++, nguyen2019multitask, Dang_2020_CVPR, Sabir2019Recurrent, nguyen2018capsuleforensics}. For the latter case, \glspl{cnn} represent the most used technology, sometimes combined with \glspl{rnn} to take advantages of the temporal information present in videos \cite{korshunov2018deepfakes, guera2018deepfake, li2018ictu, Agarwal2019protecting, Sabir2019Recurrent}.

Most deepfake detectors simply return a score indicating whether the video under analysis has been altered, not providing any explicit explanations for their predictions.
Only a few \gls{cnn}-based deepfake detectors \cite{Matern2019exploiting, nguyen2019multitask, Dang_2020_CVPR, li2019face} also return a localisation mask indicating the predicted fake areas of the image with the goal of providing some sort of explanation behind the \gls{cnn} decision process.

As an example, Malolan \etal \cite{9092227} apply \gls{lime} \cite{ribeiro2016i} and LRP \cite{10.1371/journal.pone.0130140} to deepfake detection with an Xception model, showing how the network correctly focus on the face area to compute the prediction.
Dang \etal \cite{Dang_2020_CVPR} propose a self attention module that can be added to existing detection models, it can be trained in a supervised or unsupervised paradigms, the former leading to better performances. They achieve explainability by generating attention maps that highlight localised forgeries.
Bonettini \etal in their study \cite{bonettini2020video} on ensemble of different trained \gls{cnn} models propose to use an attention layer. This produces a mask that shows the face areas that mostly impact on the classification output.
Chen and Yang \cite{chen2020manipulated} combine frequency domain analysis with semantic segmentation of the face and an attention mechanism.

In a survey, Sohrawardi \etal \cite{sohrawardidefaking} found that explainability is a key requirement in journalism for trusting a deepfake detector, they also propose a user friendly system that suggests which faces of a video are most likely fake. %

The problem of explaining deep neural networks has been tackled by several works. Buhrmester \etal widely talk about the state of the art black-box explainers in their survey \cite{buhrmester2019analysis}.
\gls{lime} \cite{ribeiro2016i} is a model agnostic explanation technique that learns an interpretable model locally around the prediction.
\gls{shap} \cite{NIPS2017_7062} is an explanation framework that assigns each feature an importance value for a particular prediction.
\gls{gradcam} \cite{8237336} is a technique for producing ``visual explanations'' for \gls{cnn}-based models.
In their work, Jetley \etal \cite{jetley2018learn} propose a multi-layer attention mechanism that is able to improve the classification performance of a network and to produce multiple attention maps at different granularity.

A non-trivial question is how to evaluate explanations, some works provide qualitative \cite{smilkov2017smoothgrad} and quantitative \cite{Santhanam2019OnEE} measures for the quality of a visual explainers, enabling the usage of sound criteria for evaluation.

In this paper we want to get deeper insights on which characteristics of a fake face are primarily considered by a classification algorithm and how different explanation techniques help achieving this goal. To this purpose we compare different techniques focusing on their application to the deepfake detection domain.

\section{Approach}
\label{sec:approach}
In order to investigate deepfake detection explainability, we consider two main actors: (i) a deepfake detector, which we consider based on \glspl{cnn} motivated by the vast majority of state of the art solutions; (ii) an explanation technique, which can be independent or not from the detector.
In this section we describe the main \gls{cnn} backbone used as detector, as well as all the used explanation methods.
We also provide some additional details about the way we specifically adapt \gls{shap} to the case of video analysis.

\subsection{Detection model employed}
\label{sec:detection_techniques}

We use EfficientNet network \cite{Tan2019EfficientNetRM} as backbone, as it proved to be powerful and lightweight at the same time. Indeed, the same architecture has also been used by the winner of the \gls{dfdc} competition \cite{selimsef}.

\begin{figure}
    \centering
    \begin{subfigure}[b]{\linewidth}
        \centering
        \includegraphics[height=\linewidth,angle=90]{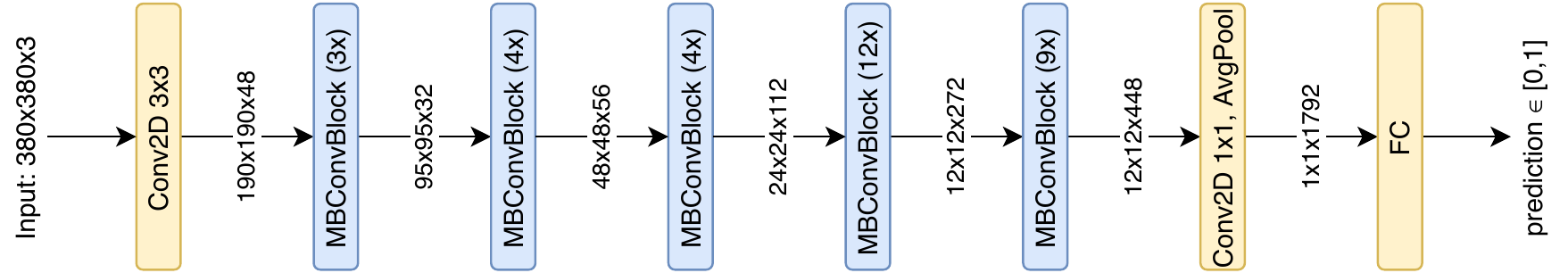}
        \caption{EfficientNet B4}
    \end{subfigure}
    
    \begin{subfigure}[b]{\linewidth}
        \centering
        \includegraphics[height=\linewidth,angle=90]{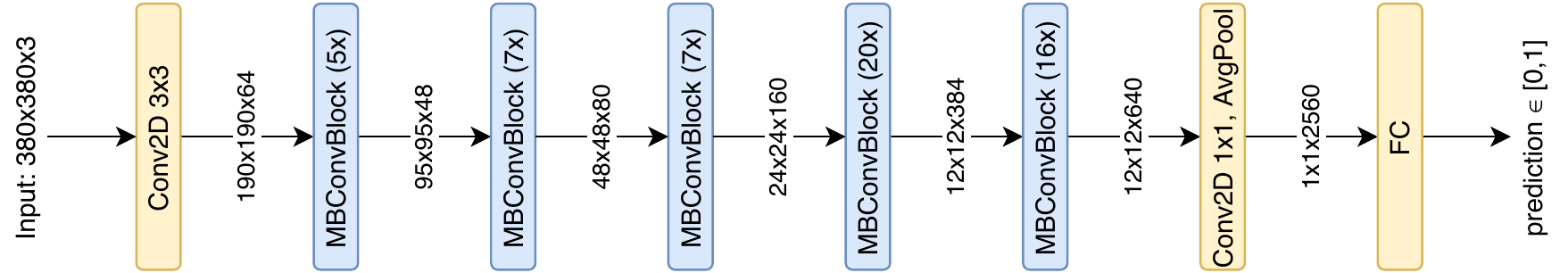}
        \caption{EfficientNet B7}
    \end{subfigure}
    \caption{The architecture of two used EfficientNet versions adapted for binary classification.}
    \label{fig:effnet_b4_b7}
\end{figure}

We train the network as binary video deepfake detector working on a per-frame basis and later use it as source model for getting explanations through different techniques. The EfficientNet architecture is built to be usable with different input resolution (224$\times$224 and 380$\times$380 in our experiments). \Cref{fig:effnet_b4_b7} shows the details of EfficientNet B4 and B7 architectures, the two versions we use. The central blocks are mobile inverted bottlenecks MBConv \cite{8578572} with squeeze-and-excitation optimisation \cite{8578843}.

\subsection{Explanation Techniques Investigated}
\label{sec:explanation_techniques}

Buhrmester \etal \cite{buhrmester2019analysis} discuss a wide variety of explanation methods, providing several formal categorisations for them. In our work, we consider three classes of explainers:
\begin{enumerate}[noitemsep]
    \item \textbf{Black-box approaches}: external tools that analyse the relation between input provided to the classification model and the resulting output. The nature of the model or its internal mechanics are not relevant to the tool, which only needs input-output samples. The explanation consists in highlighting which parts of the input are likely to be more important for causing the observed output.
    \item \textbf{White-box approaches}: external tools that take advantage from knowing the internal structure of a classification model. For neural networks they usually analyse the gradient function, providing neuron activation maps as explanations.
    \item \textbf{Embedded approaches}: models that are designed to return not only the prediction for the given input, but also an explanation for it. This is the case for \glspl{cnn} that implement an attention mechanism, since they are able to return an ``attention map'' that highlights the parts of the image on which the network focused more while taking its classification decision.
\end{enumerate}
In this work we compare specific and different explanation techniques in order to cover all of these categories.

\subsection{Black-box: SHAP with 3D segmentation}

\begin{figure}
    \centering
    \includegraphics[width=\linewidth]{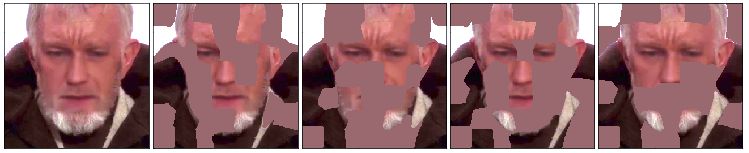}
    \caption{On the left the original image (from Mesonet dataset \cite{afchar2018mesonet}), then 4 examples of superpixel muting by Kernel SHAP.}
    \label{fig:shap_muting}
\end{figure}

\begin{figure}
    \centering
    \includegraphics[width=.3\linewidth, trim=350 70 60 60, clip]{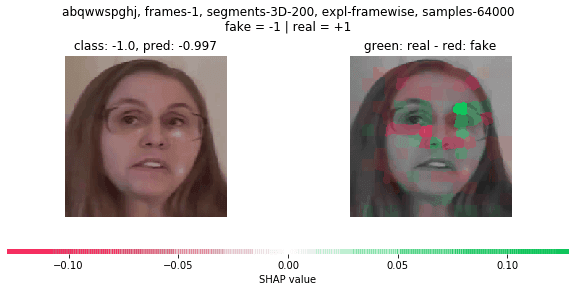}
    \caption{Example output from Kernel SHAP: superpixels are assigned different SHAP values. Negative values (negative class) are red and positive values (positive class) are green.}
    \label{fig:shap_values}
\end{figure}

For the black-box category we focus on \gls{shap} Kernel Explainer \cite{NIPS2017_7062}. It provides a versatile model-agnostic algorithm that can be easily extended to images classification. The input image is segmented into non-intersecting ``superpixels'', following the criteria of colour and spatial similarity. Several versions of the image are then created, where different superpixels are ``muted'' every time. In a nutshell, ``muting'' a superpixel means replacing its pixels with a solid colour equal to the mean of the original image, as shown in \Cref{fig:shap_muting}.
The model to be explained is then run for each modified version of the image and, based on the differences in the output with the original image, every superpixel is assigned a SHAP value describing the relation between that superpixel and the output class. A sample output is shown in \Cref{fig:shap_values}.

Since the considered deepfakes are videos and not images, we propose a modified way of segmenting frames into superpixels for \gls{shap} analysis. Image segmentation step generates what can be consider 2D superpixels, their natural extension for videos would be 3D superpixels that aggregate pixels that are both spatially and temporally close. If 2D superpixels are areas in an image, 3D superpixels can be considered as volumes in a video.
\begin{figure}
    \centering
    \begin{subfigure}[b]{.19\linewidth}
        \includegraphics[width=\linewidth]{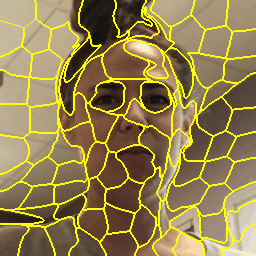}
    \end{subfigure}
    \begin{subfigure}[b]{.19\linewidth}
        \includegraphics[width=\linewidth]{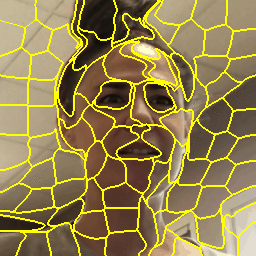}
    \end{subfigure}
    \begin{subfigure}[b]{.19\linewidth}
        \includegraphics[width=\linewidth]{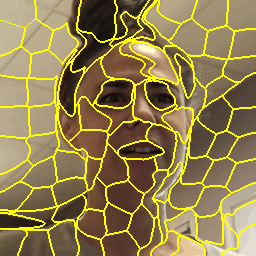}
    \end{subfigure}
    \begin{subfigure}[b]{.19\linewidth}
        \includegraphics[width=\linewidth]{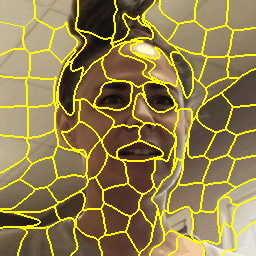}
    \end{subfigure}
    \begin{subfigure}[b]{.19\linewidth}
        \includegraphics[width=\linewidth]{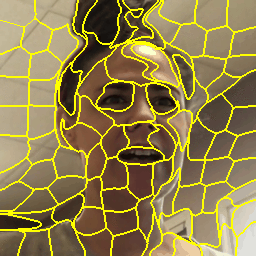}
    \end{subfigure}
    
    \begin{subfigure}[b]{.19\linewidth}
        \includegraphics[width=\linewidth]{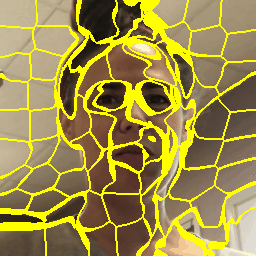}
    \end{subfigure}
    \begin{subfigure}[b]{.19\linewidth}
        \includegraphics[width=\linewidth]{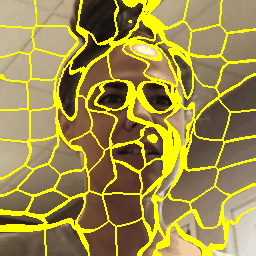}
    \end{subfigure}
    \begin{subfigure}[b]{.19\linewidth}
        \includegraphics[width=\linewidth]{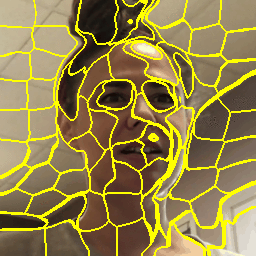}
    \end{subfigure}
    \begin{subfigure}[b]{.19\linewidth}
        \includegraphics[width=\linewidth]{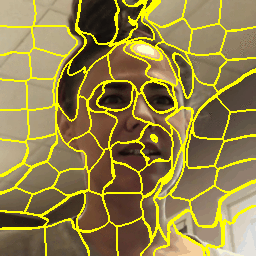}
    \end{subfigure}
    \begin{subfigure}[b]{.19\linewidth}
        \includegraphics[width=\linewidth]{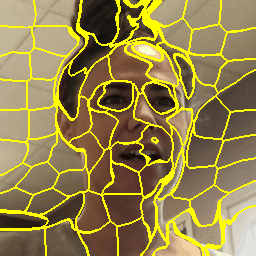}
    \end{subfigure}
    \caption{Some consecutive frames from 2D segmentation process (top) and 3D segmentation process (bottom). 2D segmentation is less consistent across consecutive frames (e.g., in the nose area).}
    \label{fig:segmentation_2d_3d}
\end{figure}
Advantages are:
\begin{itemize}[noitemsep]
    \item Segments look consistent across frames, approximately keeping their shapes also across distant frames;
    \item Volumes are identifiable as entities throughout the entire video, allowing cross-frame operations on derived 2D segments slices.
\end{itemize}
A visual comparison between 2D (frame-wise) and 3D (video-wise) segmentation is shown in \Cref{fig:segmentation_2d_3d}, where 5 consecutive frames from a video have gone through both segmentation methods.

\begin{figure}
    \centering
    \includegraphics[width=.7\linewidth]{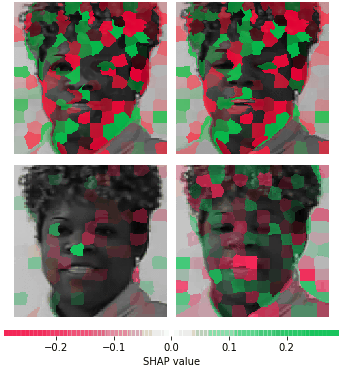}
    \caption{Two consecutive frames of a sequence explained by \gls{shap} with 3D segments in a video-wise (top) and frame-wise (bottom) fashion. Explanations were generated for the Bonettini classifier \cite{bonettini2020video} with a number of SHAP samples equal to 1000.}
    \label{fig:seg-3D_expl-videowise-framewise}
\end{figure}

Pairing a video-wise segmentation with a frame-wise classifier allows us to choose different ways to perform explanation on the segments. In particular, SHAP values can be assigned in the following ways:
\begin{itemize}[noitemsep]
    \item A single SHAP value per volume segment (video-wise explanation);
    \item A SHAP value per segment in a frame, obtaining as many independent SHAP values per volume as the number of frames in which that volume appears (frame-wise explanation).
\end{itemize}
An example is shown in \Cref{fig:seg-3D_expl-videowise-framewise}, where we show how video-wise explanation keeps the same values in consecutive frames. The video-wise explanation is necessary when the classifying model only takes videos as input (for example \glspl{rnn}). Anyway, if the classifying model accepts single frames as input, then the frame-wise explanation has the advantage to let us perform some operations on the frame-level \gls{shap} values, such as averaging values belonging to the same 3D superpixel or computing their variance. We will use this characteristic to study the \gls{shap} inter-frame consistency in \Cref{sec:intrinsic_evaluation}.

\subsection{White-box: GradCAM with binary labels}

\gls{gradcam} \cite{8237336} is a technique for producing ``visual explanations'' that uses the neural network gradients flowing from a class output neuron up to the final (or arbitrary) convolutional layer to produce a low-resolution activation map. This map highlights the spatial regions in the image that were critical for predicting that class.

\begin{figure}
    \centering
    \includegraphics[width=.6\linewidth]{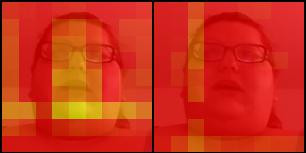}
    \caption{Output from our \gls{gradcam} extension, high values (far from zero) are yellow, low values (close to zero) are red. On the left the parts supporting prediction ``fake'', on the right the parts supporting prediction ``real''. The pixel smoothing from the original version of \gls{gradcam} has also been removed to have a clear vision of the pixel values.}
    \label{fig:gc_compl}
\end{figure}

In the original implementation, a ReLU layer is used during the creation of the activation map in order to highlight only the locations that positively contribute to the output class. This is suitable for a multi class problem where the output is one-hot encoded. Since deepfake detection, as a binary classification problem, only needs one output neuron to encode two opposite classes, we drop the ReLU layer obtaining an image that contains both positive and negative values for the pixels. We then produce 2 separate images representing the same information in absolute values. In this way we are able to see at the same time localised contributions to both classes. An example is shown in \Cref{fig:gc_compl}.

For our experiments we use the last MBConv layer of the EfficientNet model since it contains the most semantic information. When the input size is 380$\times$380, we obtain an activation map of resolution 12$\times$12. %

\subsection{Embedded: EfficientNet with self-attention}

\subsubsection{Learn to pay attention (LTPA)}

Since EfficientNet does not originally embed self-explaining mechanisms, we derived a new model by combining EfficientNet architecture with the \gls{ltpa} attention mechanism described in \cite{jetley2018learn}.

We created two different models, depending on the selected EfficientNet version: (i) EfficientNet B4 LTPA; (ii) EfficientNet B7 LTPA.

As in the original \gls{ltpa} formulation, we work with 3 attention layers. We chose the layers following some simple criteria: (i) the deepest attention layer must coincide with the final MBConv layer, as recommended in the original paper; (ii) the previous layers will be the deepest MBConv layers having different spatial resolutions.

\begin{figure}
    \centering
    \includegraphics[width=\linewidth]{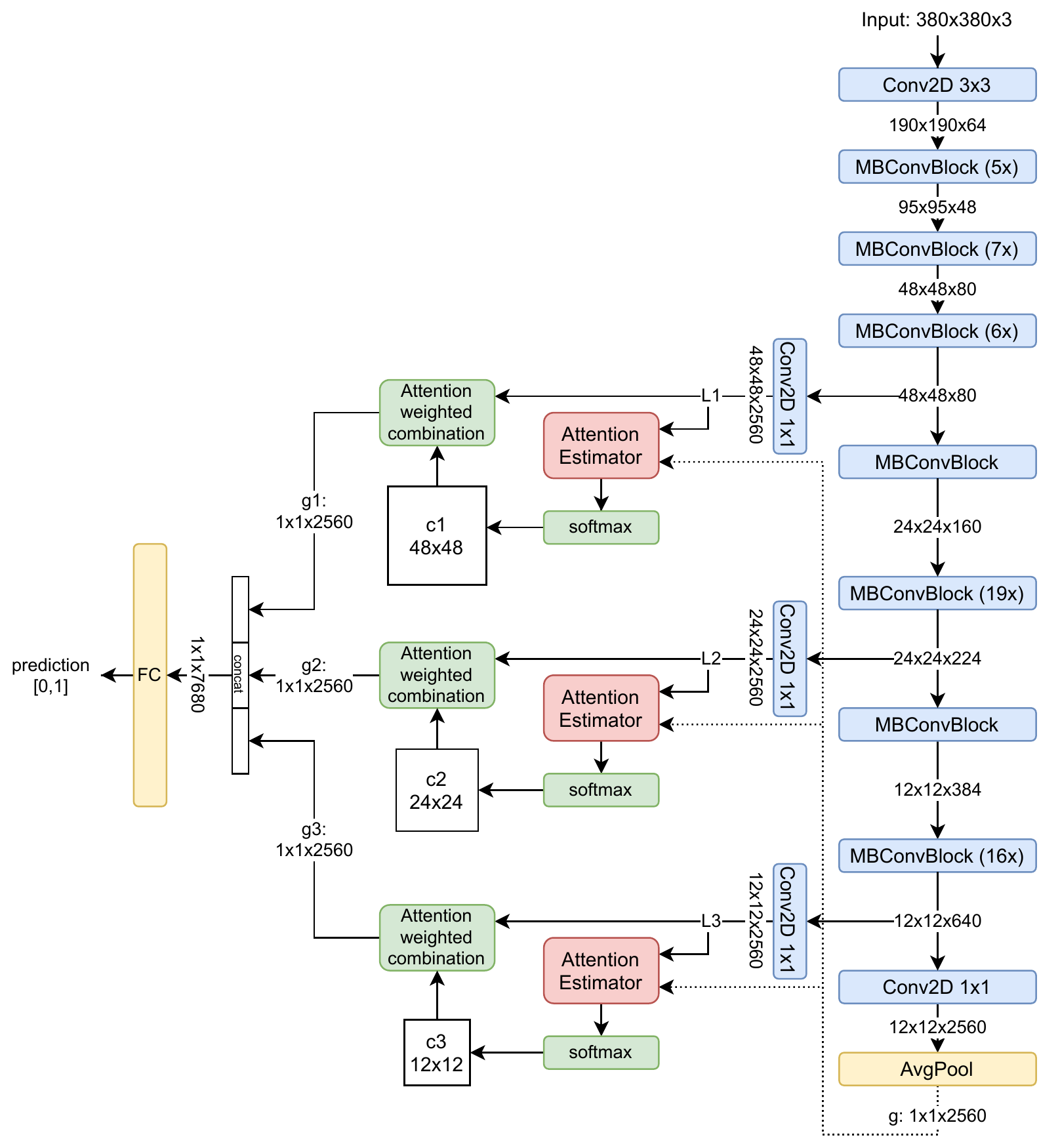}
    \caption{We combined EfficientNet B7 with the self attention mechanism by Jetley \etal \cite{jetley2018learn}, obtaining EfficientNet B7 LTPA. The network outputs a prediction (number between 0 and 1) and three attention maps (c1, c2, c3).}
    \label{fig:effnet_ltpa}
\end{figure}

In the case of EfficientNet B4 LTPA with input size 380$\times$380, the total number of convolutional blocks is 32 and the 3 attention layers are located at the end of the 10th, 22nd and 32nd layers, with resolutions 48, 24, 12. In the case of EfficientNet B7 LTPA with same input size, the 3 attention layers are located at the end of the 18th, 38th and 55th layers, outputting the same maps resolution. A scheme of the network is presented in \Cref{fig:effnet_ltpa}.

\subsubsection{Bonettini's self-attention}
We also consider the model proposed by Bonettini \etal \cite{bonettini2020video}. This adds an attention layer to EfficientNet trained as binary classifier. The model features one single attention map obtained from a central layer in the convolutional part of the network. The output map is a grey-scale image of resolution 28$\times$28 highlighting pixels that are most useful for correct classification.

\section{Experiments}
\label{sec:experiments}
In this section we report the details about the conducted experiments.
We introduce the considered setup, provide an insight about the used evaluation methodology, and comment on the obtained results.

\subsection{Experimental Setup}

\subsubsection{Dataset}

To train and test the considered classifiers we use the \gls{dfdc} dataset \cite{dolhansky2020deepfake}. At the time of experiments, only the training split was publicly available, therefore we only used this part of the dataset and subsequently divided it into train, validation and test splits for our purposes. Nevertheless, even with only one part of the dataset, we could still count on 119,154 ten seconds clips and 486 unique subjects, most of them with a resolution of 1920$\times$1080. 

The dataset portion we used is not balanced since fake videos represent about 83.9\% of the total videos. For training, we balanced the dataset following the strategy of \cite{bonettini2020video}.

\subsubsection{Model training}

In order to maximise classification accuracy, every frame is cropped to a square around the face of the actor. The face location in the frame is detected using BlazeFace \cite{bazarevsky2019blazeface}. No additional margin was added before performing the crop, as a narrower crop on the face leads to better classification accuracy \cite{9092227}. A proportional scaling (up or down) is performed on the input image when needed to match the input resolution of the networks. We use only 32 evenly-spaced frames per video as it seems enough to avoid over-fitting, while choosing more than that number would not improve the validation loss \cite{bonettini2020video}.

During the training and validation phases we used the same data augmentation policy and training parameters described by Bonettini \etal \cite{bonettini2020video}: we use Pytorch \cite{paszke2019pytorch} as deep learning framework, we perform random augmentations on the input (downscaling, horizontal flipping, brightness and colour tuning, noise and JPEG compression) with Albumentation \cite{buslaev2020albumentations}, and we use Adam \cite{kingma2014adam} optimizer with $\beta_1 = 0.9$, $\beta_2 = 0.999$, $\epsilon = 10^{-8}$, and initial learning rate equal to $10^{-5}$.

Given the dataset size, a validation loss plateau was often reached far before the end of the epoch. We use batches of 12 faces (6 real and 6 fake) randomly chosen in the train split. We stop at a maximum of 75000 iterations. Validation is performed every 1500 iterations, on 24000 samples randomly chosen in the validation split. The initial learning rate is reduced of a 0.1 factor if the validation loss does not decrease after 10 validations (15000 iterations).

We perform training combining 2 models (EfficientNet B4 and B7) and input sizes (224$\times$224 and 380$\times$380), starting from weights obtained after pre-training on ImageNet.

As an ablation study for the LTPA modules, equivalent baseline models without the attention module have been trained using the same training settings. Both the versions have been tested on the test split. The results are summarised at video level in \Cref{tab:effnet_test_video}.

\begin{table}
        \caption{The test results at video level (4328 real and 20460 fake videos).}
    \label{tab:effnet_test_video}
    \centering
    \footnotesize
    \begin{tabular}{l|c|c}
        \toprule
        EfficientNet Model       & Bal. accuracy   & ROC AUC \\
        \midrule
        B4, 224$\times$224       & 0.888                 & 0.958   \\
        \textbf{B4, 380$\times$380}       & \textbf{0.931}                 & \textbf{0.980}   \\
        B7, 224$\times$224       & 0.906                 & 0.964   \\
        \textbf{B7, 380$\times$380}       & \textbf{0.926}                 & \textbf{0.976}   \\
        \hline
        B4, LTPA, 224$\times$224 & 0.879                 & 0.946   \\
        \textbf{B4, LTPA, 380$\times$380} & \textbf{0.929}                 & \textbf{0.978}   \\
        B7, LTPA, 224$\times$224 & 0.893                 & 0.962   \\
        B7, LTPA, 380$\times$380 & 0.904                 & 0.972   \\
        \bottomrule
    \end{tabular}
\end{table}

\subsubsection{Explanations generation}

\begin{figure}
    \centering
    \includegraphics[width=\linewidth]{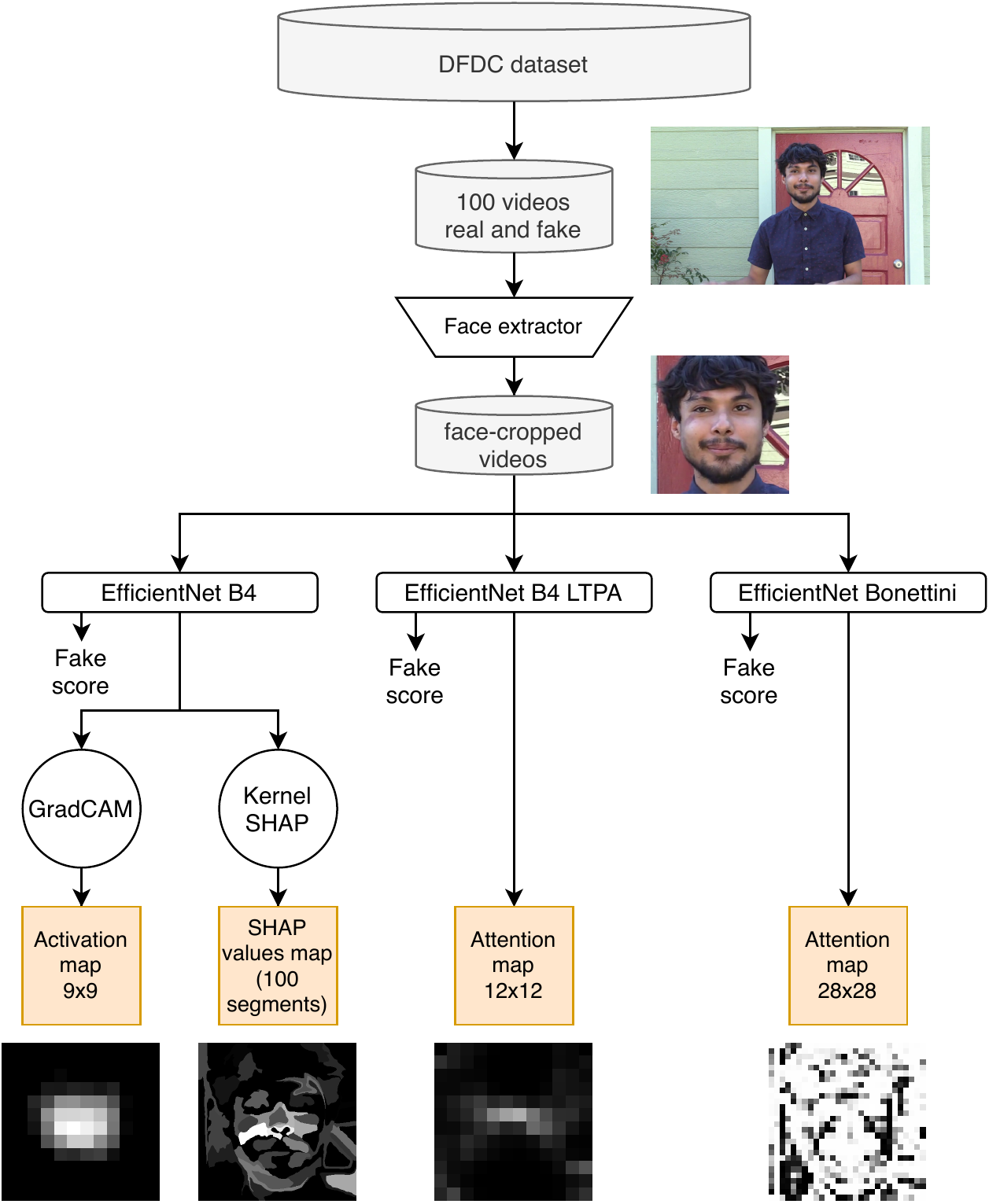}
    \caption{The steps to generate predictions and explanations from the dataset videos.}
    \label{fig:expl_pipeline}
\end{figure}

The experiment pipeline is shown in \Cref{fig:expl_pipeline}. We chose 100 videos to be explained, randomly sampled in the test split. For each video we extract 100 frames evenly spaced across the video, resulting in sequences 10 frames per second. For each frame we extract a square containing only the face and resize it to a fixed size so that the crop size is uniform across the sequence.

For each video, explanations are generated using (i) EfficientNet B4 LTPA 380$\times$380 (ii) EfficientNet B4 380$\times$380 + GradCAM (iii) EfficientNet B4 380$\times$380 + Kernel SHAP with 2000 samples (iv) attention mask from \cite{bonettini2020video}.

For each frame explanation, a gray-scale image is produced. For LTPA and Bonettini, the explanation is the output attention mask as is. For GradCAM and SHAP we choose the masks that activate for fake inputs. GradCAM and SHAP explanations must be normalised before being converted to a gray-scale image. In particular, we linearly normalise their values so that the extreme pixel (the furthest from zero, in the positive or negative direction) has absolute value equal to 1.

\subsection{Evaluation methodology}
\label{sec:evaluation_criteria}

We selected from the dataset test split a subset of videos that fulfil the following requirement: the mean derivative of the detected face bounding box edges, normalised to the frame resolution, should be smaller than 0.002. This selects only videos for which the face detector is stable on the position of the face, filtering out the videos for which the face detector was performing worst, jumping wildly to places that do not contain any face. In this way we selected a pool of 58 fake videos.

Some evaluation metrics have been investigated for visual explainability \cite{Petsiuk2018rise}. However, they may require a clean ground truth, or to slightly perturb the input image before metric computation.
In the considered forensic scenario, we only have a fuzzy ground truth, and we cannot afford editing the input image, as manipulation is what we want to detect.
For this reason, we propose an alternative set of evaluation criteria that are simple and reasonable.

\subsubsection{Variance}

We define variance $V$ of an explainer as the average pixel variance within individual frames:
\begin{equation}
    V = \operatorname{avg}_{videos} \left( \operatorname{avg}_{f \in frames} \left( \operatorname{var}(f) \right) \right)
\end{equation}
where we average over our pool of fake videos. We note that higher values for $V$ could indicate (i) more specific features being identified (a positive property), or (ii) more noise in the mask (a negative property).

\subsubsection{Inter-frame consistency}

To evaluate the inter-frame consistency, we compute the Pearson Correlation Coefficient between consecutive frames $x$ and $y$:
\begin{equation}
    \operatorname{PCC}_{x,y} = \frac{\sum_i{(x_i - \bar{x})(y_i - \bar{y})}}{ \sqrt{\sum_i{(x_i - \bar{x})^2}} \sqrt{\sum_i{(y_i - \bar{y})^2}}}
\end{equation}
where $\bar{x}$ and $\bar{y}$ are the mean values of the frames. We then average this quantity over all the frames and over all the videos in our pool, defining inter-frame consistency $\tau$ as:
\begin{equation}
    \tau = \operatorname{avg}_{videos} \left( \operatorname{avg}_{f \in frames} \left( \operatorname{PCC}_{f,f+1} \right) \right)
\end{equation}
The value of $\tau$ will be in a range between -1 and 1, with higher values indicating better performance.

\subsubsection{Intra-frame consistency}

We consider an explanation frame intra-consistent if the highlighted pixels are in a correlation with the neighbouring ones. Let $a_{i,j}(x)$ be the auto-correlation coefficients of an explanation frame $x$ computed for the vertical lag $i$ and the horizontal lag $j$:
\begin{equation}
    a_{i,j}(x) = \sum_{h,w} { (x_{h,w} - \bar{x}) (x_{h+i,w+j} - \bar{x}) }
\end{equation}
We define the intra-frame consistency $\rho$ as the correlation between the image and itself shifted 10\% in 4 directions (up, down, left, right).
Formally, $\rho$ is the mean value of $a_{i,j}(x)$ for $(i,j) \in S$, normalised by the factor $a_{0,0}(x)$:
\begin{equation}
    \rho(x) = \frac{ \operatorname{avg}_{(i, j) \in S} ( a_{i, j}(x) ) }{ a_{0,0}(x) }.
\end{equation}
where $S = \lbrace (0, \frac{l}{10}), (0,-\frac{l}{10}), (\frac{l}{10}, 0), (-\frac{l}{10},0) \rbrace $ is a set of vertical and horizontal shifts of 10\% of the image size $l$. Note that the $\rho$ computed on images of random noise tend to be close to 0. Similarly to the previous metrics, we compute a final $\rho$ as an average over all frames and all videos. Values lie in the range [-1,1] with the higher value the better.

\subsubsection{Centredness}

The area of the face that is subject to deepfake manipulation is only a fraction of the entire input, and always centred, from the way we are extracting faces. We expect a good explainer to mostly highlight pixels on the face, therefore we define centredness $\mu$ as:
\begin{equation}
    \mu = \operatorname{avg}_{videos} \left( \operatorname{avg}_{f \in frames} \left( \frac{ \sum_{i \in \text{inner 50\%}}{f_i^2} }{ \sum_{i \in f}{f_i^2} } \right) \right)
\end{equation}
where $f_i$ is the $i$-th pixel of $f$ and ``inner 50\%'' is a centred square crop having half the area of the full frame. In our evaluation, the higher $\mu$ the better.

\subsubsection{Subjective evaluation (user study)}

\begin{figure}
    \centering
    \begin{subfigure}[b]{0.3\linewidth}
        \centering
        \includegraphics[width=\linewidth, trim=20 20 20 20, clip]{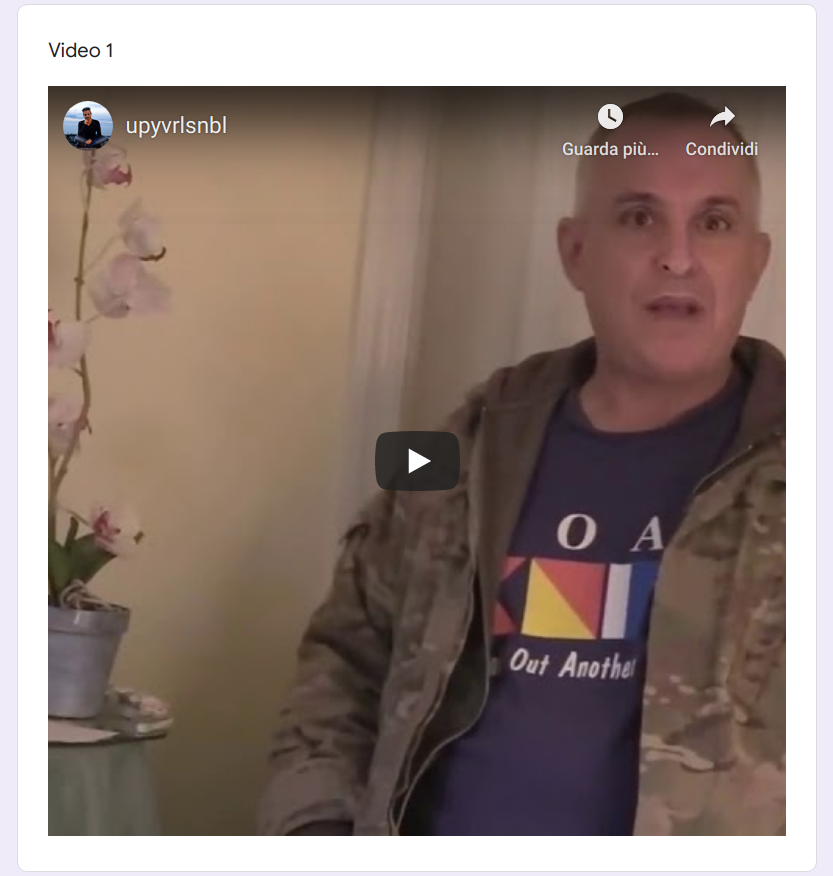}
    \end{subfigure}
    \hfill
    \begin{subfigure}[b]{0.3\linewidth}
        \centering
        \includegraphics[width=\linewidth, trim=20 20 20 20, clip]{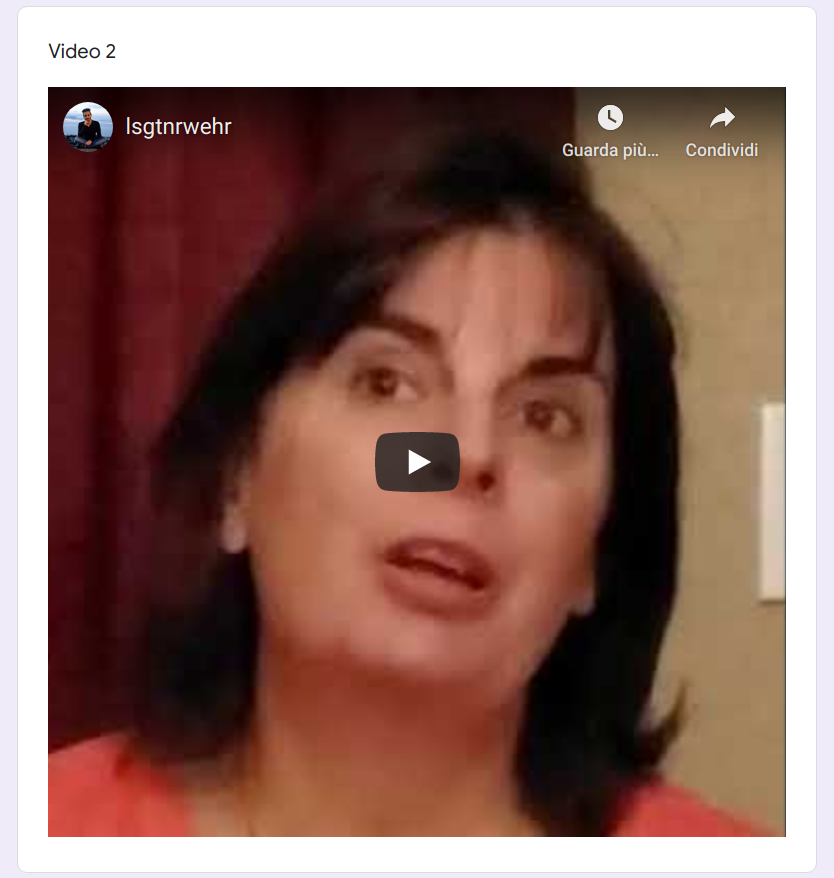}
    \end{subfigure}
    \begin{subfigure}[b]{0.38\linewidth}
        \centering
        \includegraphics[width=\linewidth, trim=40 20 450 20, clip]{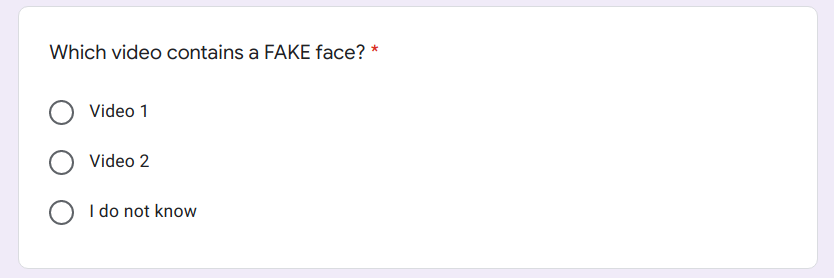}
    \end{subfigure}
    \caption{The first question of a survey section.}
    \label{fig:survey_q1}
\end{figure}

\begin{figure}
    \centering
    \includegraphics[width=\linewidth]{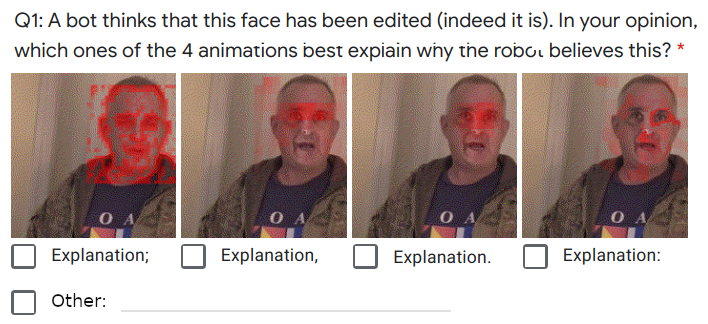}
    \caption{The second question of a survey section.}
    \label{fig:survey_q2}
\end{figure}

We perform a user study to determine which explainer is preferred at visual inspection. In literature, other surveys have been performed as well to evaluate explainers \cite{8237336}. We prepare a survey of 20 sections, each consisting of the following 2 questions:
\begin{enumerate}[noitemsep]
    \item Two videos are presented to the user, one real and one fake, the user is asked to watch them and determine which is the fake one, as show in \Cref{fig:survey_q1}. This first question has the goal to make the user focus on the video details and individuate possible hints for a video being fake, as a preparation for the next question; it also gives us an idea of the users' accuracy.
    \item After the user answered the previous question, the correct answer is revealed; we show again the fake video, together with 4 explanations that highlight why the classifiers would predict the video as fake; the user is asked to pick zero or more explanations he/she agrees with, as shown in \Cref{fig:survey_q2}.
\end{enumerate}

The order of the answer options is randomised. Given the number of videos, the survey has been divided into two halves having 10 questions each. We propose one of the two halves to each participant. 

\begin{figure}
    \centering
    \begin{subfigure}[b]{.24\linewidth}
        \includegraphics[width=\linewidth]{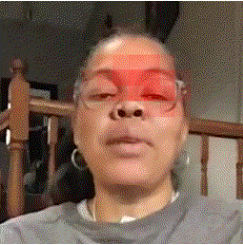}
        \caption{GradCAM}
        \label{fig:expl_gc}
    \end{subfigure}
    \begin{subfigure}[b]{.24\linewidth}
        \includegraphics[width=\linewidth]{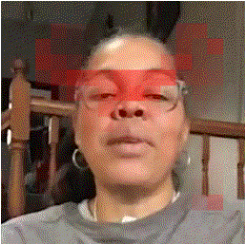}
        \caption{LTPA lv. 2}
        \label{fig:expl_ltpa2}
    \end{subfigure}
    \begin{subfigure}[b]{.24\linewidth}
        \includegraphics[width=\linewidth]{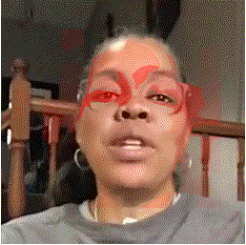}
        \caption{SHAP}
        \label{fig:expl_shap}
    \end{subfigure}
    \begin{subfigure}[b]{.24\linewidth}
        \includegraphics[width=\linewidth]{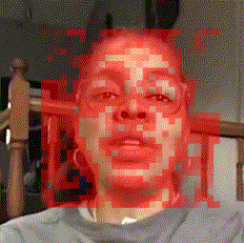}
        \caption{Bonettini}
        \label{fig:expl_bon}
    \end{subfigure}
    \caption{Explanations on the same frame of a fake video.}
    \label{fig:expl_examples}
\end{figure}

In the second question, 5 options are presented: four explanations (shuffled for each user) and an ``Other'' option with a free text field (invariably at the bottom). To present the different explanations in an homogeneous way, we use a red overlay on the subject face, assigning a transparent colour to the lowest value pixels in the explanation, and full red colour for the highest value pixels in the explanation. All the explanations are normalised to have their max value equal to 1 before being overlaid on the images. Examples of generated explanations are shown in \Cref{fig:expl_examples}.

We randomly chose 20 fake videos from the pool of 58 videos, and 20 real videos from the train and validation splits. All videos contain only one actor each and the same actor is never in more than one video.
Among the 20 fake videos, 19 are predicted as fake with high confidence (score $>$ 0.85) by at least one of the classifiers. %

\subsection{Comparison based on Intrinsic Evaluation}
\label{sec:intrinsic_evaluation}

We compute the metrics presented in \Cref{sec:evaluation_criteria} for all the presented explanation techniques, on the 58 fake videos pool. The results are summarised in \Cref{tab:metrics_evaluation}.

Some explainers, like SHAP and our GradCAM extension produce 2 maps: one to support the real prediction and another to support the fake prediction. Since we are using fake videos, we will only take into consideration the maps supporting fake prediction. The LTPA explainer will produce 3 maps at different resolutions. Given that all the 3 maps are used for the final prediction, we decided to aggregate them into a single map, computed by up-scaling all the maps to the highest-resolution one and then summing them, finally we do a max value normalisation to bring the pixel values back in the range from 0 to 1.

\begin{table}
    \caption{The average values of the considered metrics for different explainers. The value range for each metric is also shown in square brackets. For the last 3 columns, the higher the better.}
    \label{tab:metrics_evaluation}
    \centering
    \footnotesize
    \begin{tabular}{l|c|c|c|c}
        \toprule
        & $V$ &$\tau$ &$\rho$ &$\mu$ \\
        &[0, 1] &[-1, 1] &[-1, 1] &[0, 1] \\
        \midrule
        Bonettini & 0.0951 & 0.7390 & 0.1262 & 0.5286 \\
        GradCAM   & 0.0135 & \textbf{0.8756} & \textbf{0.7489} & \textbf{0.8666} \\
        LTPA      & 0.0108 & 0.7991 & 0.3333 & 0.6386 \\
        SHAP      & 0.0302 & 0.4496 & 0.2326 & 0.7348 \\
        \bottomrule
    \end{tabular}
\end{table}

\subsection{Comparison based on User Survey}

The user study was performed on 67 subject, for a total of 670 video questions answered. On average, the users correctly identified 85\% of the fake/real couples. The preference results are summarised in \Cref{tab:user_study_results}.

\begin{table}
    \caption{Results of the user study.}
    \label{tab:user_study_results}
    \footnotesize
    \centering
    \begin{tabular}{llrr}
        \toprule
        \textbf{Number of answers} & \multicolumn{3}{c}{67} \\
        \hline
        \multirow{2}{*}{\textbf{Screen used}} & Large & 43\% & \\
        & Small & 57\% & \\
        \hline
        \multirow{3}{*}{\textbf{Are you familiar with deepfakes?}} & Yes & 37\% & \\
        & Heard of it & 33\% & \\
        & No & 30\% & \\
        \hline
        \textbf{Correct video identification} & \multicolumn{3}{c}{85\%} \\
        \hline
        \multirow{4}{*}{\textbf{Explainer choices}} & GradCAM  & 165 & 23\% \\
        & \textbf{SHAP } & \textbf{221} & \textbf{31\%} \\
        & LTPA  & 137 & 19\% \\
        & Bonettini  & 185 & 26\% \\
        \bottomrule
    \end{tabular}
\end{table}

In order to validate the statistical relevance of the survey, we perform the Sign Test on pair comparisons. To determine whether we can reject the null hypothesis for which none of the two explainers is better than the other, we compare the 2-tail $p$-value with an $\alpha$ value corrected with a Bonferroni factor. In our case $ \alpha = 0.05/6 = 0.00833 $. If the $p$-value is smaller than $\alpha$ we consider the test passed (i.e. the difference in votes between two explainers is statistically relevant), otherwise we consider the test not passed. The test results are presented in \Cref{tab:sign_test}.

\begin{table}
    \caption{Sign Test results.%
    }
    \label{tab:sign_test}
    \footnotesize
    \centering
    \begin{tabular}{lllll}
        \toprule
        Hypotesis (A$>$B) & A$>$B & B$>$A & 2-tail p-value & Passed \\
        \midrule
        SHAP $>$ Bonettini:      & 212 & 176 & 0.075456 & No \\
        SHAP $>$ GradCAM:        & 210 & 154 & 0.003881 & Yes \\
        SHAP $>$ LTPA:           & 212 & 128 & 0.000006 & Yes \\
        Bonettini $>$ GradCAM:   & 178 & 154 & 0.299944 & No \\
        Bonettini $>$ LTPA:      & 177 & 129 & 0.007115 & Yes \\
        GradCAM $>$ LTPA:        & 156 & 128 & 0.108958 & No \\
        \bottomrule
    \end{tabular}
\end{table}

\subsection{Discussion}

\begin{figure}
    \centering
    \includegraphics[width=\linewidth,trim=10 20 0 50,clip]{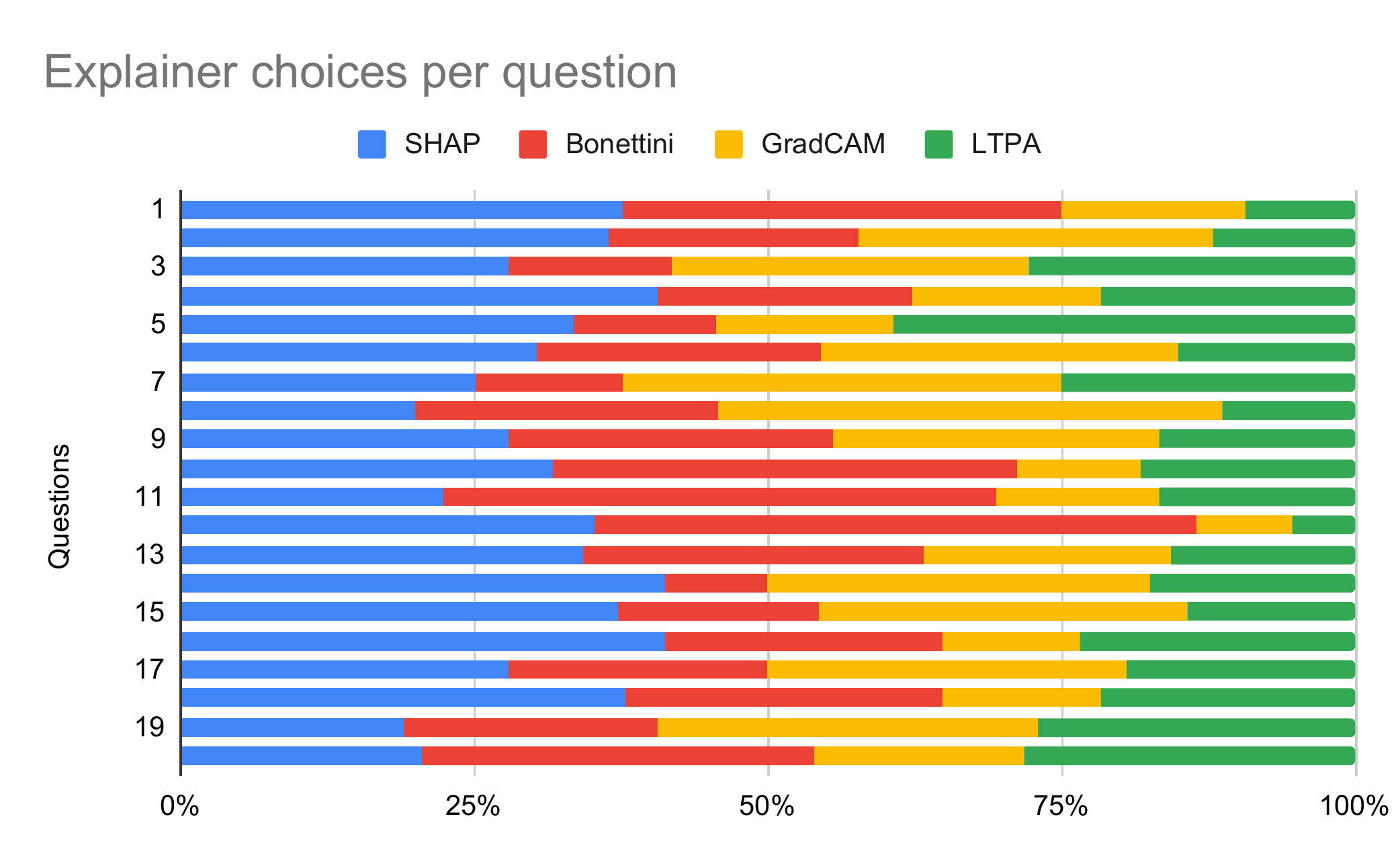}
    \caption{Explainers preferences aggregated per question.}
    \label{fig:survey_question_aggr}
\end{figure}

We noticed that the preferred explainer depended mostly on the question, i.e., the video, as shown in \Cref{fig:survey_question_aggr}.
This result may be related to the fact that the dataset from which we chose the videos to use in the survey, is actually built using different deepfake techniques, that eventually lead to different quality results of the final deepfake. %

With respect to efficiency, explainers are different. While LTPA, Bonettini and GradCAM do not bring a relevant overhead to the underlying classification model, this is not true for Kernel SHAP, that needs thousands of runs of the classification model to produce an explanation. Therefore SHAP is more expensive than the underlying model by a factor that increases linearly with the sampling number.

From the flexibility point of view, the black-box methods are by design better than the others.
GradCAM and similar white-box techniques need the model to be a convolutional network, posing limitations to their applicability. Finally, the attention methods are the least flexible, as they require the architecture to be redesigned and the model to be retrained.

\section{Conclusions}
\label{sec:conclusions}
Deepfakes have recently proved to be a major threat, thus multiple detectors have been proposed in the literature.
However, most of these methods only return a detection score, not providing any additional information on the visual cues that make a detector take a specific decision.

In this paper, we investigated methods for deepfake detectors explainability.
Given a video and a deepfake detector, we want to understand which regions of the video mostly triggered the detector decision.
To do so, we adapted, extended and compared four different explainability methodologies that provide a mask that highlights which pixels are most important to a detector.

Our experimental campaign has been conducted on the \gls{dfdc} dataset considering both quantitative and qualitative measures.
Results have shown that the human perception of a deepfake explanation mask is not always aligned with objective metrics.
Despite some methods provide explanations that are objectively better localised, other techniques tend to be preferred by many subjects.
Moreover, people opinion tends to be different from video to video.

Results suggest that a single explanation method may not be sufficient to help forensic investigators.
Future work will be devoted to study how much each explanation methodologies impact on investigators detection performance.

{\small
\bibliographystyle{ieee_fullname}
\bibliography{references}
}

\end{document}